\documentclass[sigconf, screen, authorversion, nonacm]{acmart}
\AtBeginDocument{%
  }

\setcopyright{acmlicensed}
\copyrightyear{2025}
\acmYear{2025}
\acmDOI{XXXXXXX.XXXXXXX}
\acmConference[ACM MM]{The 33rd ACM International Conference on Multimedia}{October 27--31, 2025}{Dublin, Ireland}
\acmISBN{978-1-4503-XXXX-X/2018/06}

\usepackage{enumitem}

\DeclareMathOperator*{\argmax}{arg\,max}
\usepackage{multirow}
\usepackage{makecell}

\begin{document}

\title[MUDI: A Multimodal Biomedical Dataset for Understanding Pharmacodynamic Drug-Drug Interactions]{MUDI: A Multimodal Biomedical Dataset for Understanding~Pharmacodynamic Drug-Drug Interactions}


\author{Tung-Lam Ngo}
\authornotemark[1]
\email{22028092@vnu.edu.vn}
\orcid{0009-0002-6331-8916}
\affiliation{%
  \institution{VNU University of Engineering and Technology (VNU-UET)}
  \city{}
  \country{}
}

\author{Ba-Hoang Tran}
\authornotemark[1]
\email{21020631@vnu.edu.vn}
\orcid{0009-0001-7796-142X}
\affiliation{%
  \institution{VNU University of Engineering and Technology (VNU-UET)}
  \city{}
  \country{}
}

\author{Duy-Cat Can}
\authornote{Shared first authors.}
\email{duy-cat.can@chuv.ch}
\orcid{0000-0002-6861-2893}
\affiliation{%
  \institution{Lausanne University Hospital (CHUV)}
  \institution{University of Lausanne (UNIL)}
  \city{}
  \country{}
}
\additionalaffiliation{
  \institution{VNU University of Engineering and Technology (VNU-UET)}
  \city{}
  \country{}
}

\author{Trung-Hieu Do}
\email{dotrunghieu05220161@daihocyhanoi.edu.vn}
\affiliation{%
  \institution{Hanoi Medical University}
  \institution{National Geriatric Hospital, Hanoi}
  \city{}
  \country{}
}

\author{Oliver Y. Ch\'en}
\email{olivery.chen@chuv.ch}
\orcid{0000-0002-5696-3127}
\affiliation{%
  \institution{Lausanne University Hospital (CHUV)}
  \institution{and University of Lausanne (UNIL)}
  \city{}
  \country{}
}

\author{Hoang-Quynh Le}
\authornote{Corresponding authors.}
\email{lhquynh@vnu.edu.vn}
\orcid{0000-0002-1778-0600}
\affiliation{%
  \institution{VNU University of Engineering and Technology (VNU-UET)}
  \city{}
  \country{}
\vspace{1cm}
}

\renewcommand{\shortauthors}{Ngo et al.}

\begin{abstract}
Understanding the interaction between different drugs (drug-drug interaction or DDI) is critical for ensuring patient safety and optimizing therapeutic outcomes.
Existing DDI datasets primarily focus on textual information, overlooking multimodal data that reflect complex drug mechanisms.  
In this paper, we (1) introduce MUDI, a large-scale \textbf{M}ultimodal biomedical dataset for \textbf{U}nderstanding pharmacodynamic \textbf{D}rug-drug \textbf{I}nteractions, and (2) benchmark learning methods to study it.  
In brief, MUDI provides a comprehensive multimodal representation of drugs by combining pharmacological text, chemical formulas, molecular structure graphs, and images across 310,532 annotated drug pairs labeled as Synergism, Antagonism, or New Effect.  
Crucially, to effectively evaluate machine-learning based generalization, MUDI consists of unseen drug pairs in the test set. 
We evaluate benchmark models using both late fusion voting and intermediate fusion strategies.  
All data, annotations, evaluation scripts, and baselines are released under an open research license.
\end{abstract}

\ccsdesc[500]{Information systems~Multimedia databases}
\ccsdesc[500]{Computing methodologies~Machine learning approaches}
\ccsdesc[500]{Applied computing~Life and medical sciences}
\ccsdesc[500]{Information systems~Information extraction}
\ccsdesc[300]{Computing methodologies~Supervised learning by classification}
\ccsdesc[300]{Applied computing~Bioinformatics}
\ccsdesc[300]{Applied computing~Health informatics}
\ccsdesc[300]{Computing methodologies~Neural networks}

\keywords{%
    Multimodal Dataset,
    Drug-Drug Interaction,
    Pharmacodynamics,
    Biomedical Data Mining,
    Multimodal Fusion
}
\received{30 May 2025}
\received[revised]{xx xxx 2025}
\received[accepted]{xx xxx 2025}

\maketitle

\section{Introduction}
Polypharmacy, the concurrent use of multiple medications, is common in treating complex diseases or comorbid conditions, especially in elderly patients.  
It can lead to drug-drug interactions (DDIs), where one drug alters the effect of another, potentially reducing efficacy or causing adverse events~\cite{mcquade2021drug}.
%
A promising approach is to predict DDIs proactively when assigning drugs, improving patient safety and treatment outcomes while reducing healthcare costs.  
To do so effectively requires two key components: a suitable predictive method and a well-targeted dataset for training and evaluation.


Designing such a dataset must reflect the inherently multimodal nature of DDIs, which arise from diverse pharmacological foundations, including chemical properties, pharmacological description, and molecular structure~\cite{asada2021using}.  
Yet most existing datasets focus narrowly on one of these modalities, typically textual data~\cite{zhao2016syntax, sahu2018drug, zhu2020extracting}, limiting models' ability to capture complex biochemical interactions.
Some recent multimodal DDI studies address this, but often rely on fragmented or non-standardized sources, with limited modality coverage or label inconsistency.
For example, the state-of-the-art datasets, such as DDInter~\cite{xiong2022ddinter}, simply merge pharmacokinetic and pharmacodynamic interactions into a single label set and ignore interaction directionality.
These lines of evidence suggest a need for a comprehensive, well-curated multimodal dataset.


To address these gaps, we introduce \textbf{MUDI} -- a \textbf{M}ultimodal Biomedical Dataset for \textbf{U}nderstanding Pharmacodynamic \textbf{D}rug-Drug \textbf{I}nteractions. 
MUDI is a large-scale, richly annotated collection of paired drugs, integrating multiple data modalities: drug descriptions, molecular structure graphs, molecular structure images, and chemical formulas.
Unlike existing datasets, MUDI provides directed labels (e.g., Synergism, Antagonism) and undirected labels (e.g., New Effect).  
Notably, MUDI's test set contains a substantial portion of interactions involving unseen drugs to assess model generalization.
Moreover, we provide multimodal baselines with intermediate and late fusion to benchmark learning from heterogeneous biomedical inputs.
Finally, we release the full MUDI dataset, annotation guidelines, baseline implementations, and benchmarking results publicly under an open license. 

We summarize our key contributions as follows:
\begin{itemize}[leftmargin=2em]
    \item We construct and release \textbf{MUDI}, a multimodal pharmacodynamic DDI dataset, curated based on clinical knowledge, containing $310{,}532$ annotated drug pairs.
    \item We provide benchmark results using \textbf{multimodal baselines} for predicting DDIs from heterogeneous biomedical inputs.
    \item We publicly release the dataset, annotations, code, and evaluation pipelines to support reproducible multimodal research.
\end{itemize}

\section{Related Work}

\paragraph{DDI Prediction Datasets}
Numerous datasets have been developed for DDI prediction, varying in scope and annotation strategies. 
DrugBank~\cite{wishart2018drugbank} is the most comprehensive and widely used.
However, its DDI information is presented as free-text descriptions rather than structured labels or relations.
The associated drug metadata is also not systematically organized into distinct modalities, limiting its applicability for multimodal machine learning.
Several DDI-focused datasets have been constructed based on DrugBank and supplementary sources such as FAERS~\cite{faers2024}, MEDLINE~\cite{medline2024}, and other biomedical databases. 
These datasets are typically single-modal, providing DDI information using either text or SMILES strings. 
Among them, some describe interactions without assigning labels (e.g., LIDDI~\cite{banda2015provenance}, TDC~\cite{huang2021therapeutics}, BioSNAP~\cite{biosnapnets}),
while others provide binary or inconsistently defined label sets without a systematic scheme, and often omit relation directionality (e.g., SemEval 2013~\cite{bjorne2013uturku}, HODDI~\cite{wang2025hoddi}, TWOSIDES~\cite{tatonetti2012data}, Mendeley DDI dataset~\cite{yu2020data}).

The DDInter dataset~\cite{xiong2022ddinter} represents a recent state-of-the-art with emphasis on multimodality.
However, it merges pharmacokinetic and pharmacodynamic interactions into a single label set, ignoring their distinct roles and causal links.
Its assumption of symmetric interactions overlooks real-world directionality, where one drug may affect another without reciprocal effects.
Finally, molecular graphs are not provided, missing key spatial and topological cues critical for modeling molecular structure.

\paragraph{DDI Multimodal Approaches}
Recently, there has been a growing trend towards multimodal approaches for drug-drug interaction prediction.
Most studies aim to incorporate, expand, and diversify drug-related multimodal information from various data sources to improve model performance.
However, the richness and availability of these modalities remain limited.
Some studies extract additional aspects from textual descriptions, such as targets and pathways~\cite{gan2023dmfddi, deng2020multimodal}, yet this remains multi-aspect features derived from a single modality.  
Others incorporate knowledge graphs~\cite{shi2024subge, asada2023integrating}, producing text-graph hybrids rather than fully multimodal models.  
Notably, some researchers have utilized SMILES strings from DrugBank to capture molecular structural information~\cite{mondal2020bertchem, asada2021using} or chemical substructures~\cite{deng2020multimodal}.
More advanced methods, such as 3DGT-DDI~\cite{he20223dgt}, leverage 3D molecular graph structures obtained from PubChem to better represent molecular geometry.
Nevertheless, current multimodal DDI research still relies on fragmented sources, narrow modality coverage, and non-standardized preprocessing.

\paragraph{How MUDI Complements Existing Resources}
MUDI serves as a standardized and reproducible resource for multimodal DDI prediction, addressing key limitations of prior datasets (see Table~\ref{tab:data_comparison}).  
MUDI is specifically curated for pharmacodynamic interactions, with biologically grounded labels and interaction directionality.
Unlike prior fragmented datasets, MUDI integrates six structured modalities, supporting standardized evaluation and downstream pharmacological integration.
Finally, MUDI emphasizes \textbf{accessibility and reproducibility}, releasing the full dataset, annotation guidelines, and baseline implementations under an open license to foster transparent, accelerated research in multimodal biomedical AI.

\section{Dataset Construction}

\begin{figure}[]
    \centering
    \includegraphics[width=\linewidth]{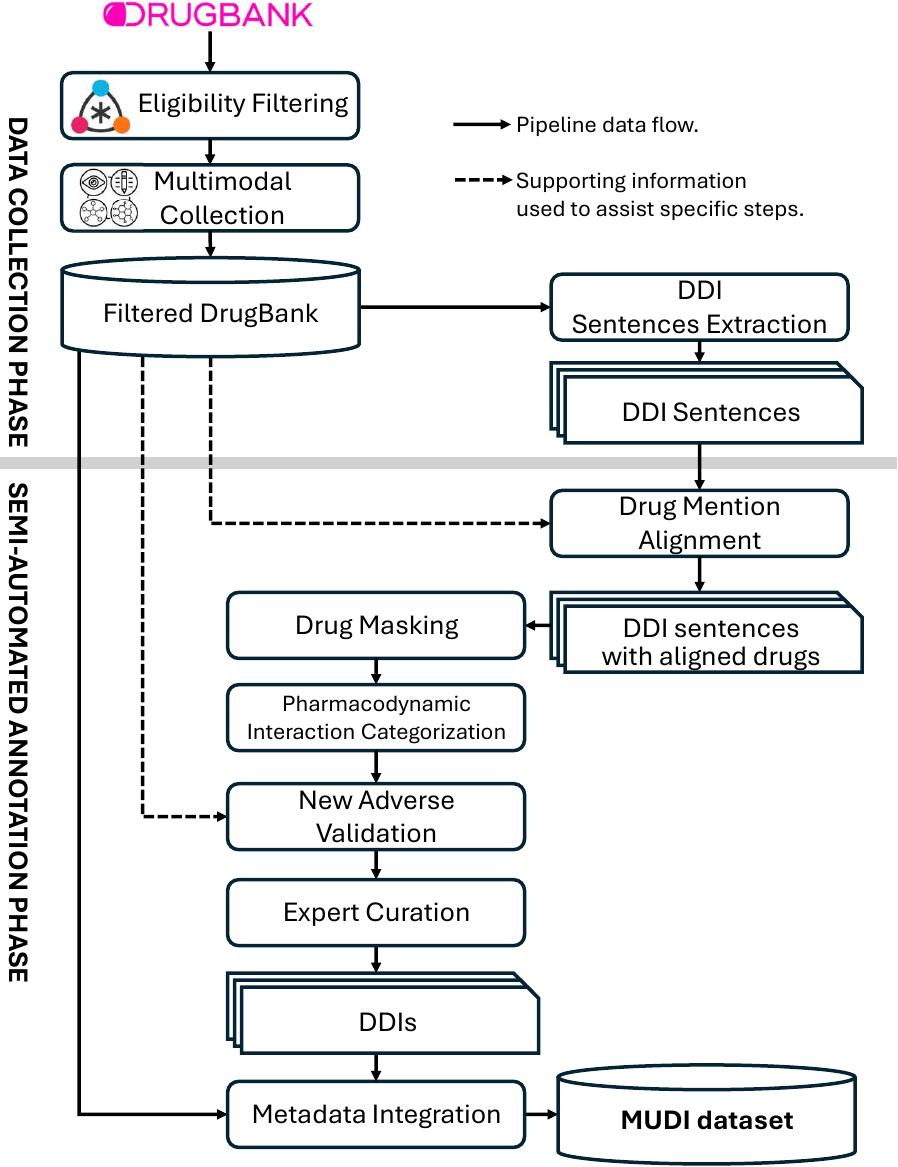}
    \caption{Overview of MUDI dataset construction pipeline.} 
    \Description{
        Raw drug data is extracted from DrugBank, followed by multimodal preprocessing to generate textual fields, molecular graphs, chemical structure images, and normalized formulas.
        Interaction labels are assigned based on pharmacodynamic relationship types.
        The resulting dataset supports multimodal DDI prediction with generalization to unseen drugs.
    }
    \label{fig:data_pipeline}
\end{figure}

MUDI is constructed from a pharmacodynamic perspective on drug-drug interactions, targeting clinically meaningful effects resulting from drug co-administration.
Each interaction is categorized into one of three abstract-level pharmacodynamic labels:
\begin{itemize}[leftmargin=2em]
    \item \textbf{\texttt{Synergism}} (directed relationship): The co-administration of two drugs results in an enhanced effect of one of the drugs.
    This enhancement may involve therapeutic efficacy, but it can also manifest as increased toxicity or adverse effects.
    \item \textbf{\texttt{Antagonism}} (directed relationship): The concurrent use of two drugs reduces or neutralizes the effect of one of the drugs.
    This reduction may pertain to therapeutic benefit, but may also involve diminished toxicity or side effects.
    \item \textbf{\texttt{New Effect}} (undirected relationship): The combination of two drugs leads to a new effect -- either adverse or therapeutic -- that is not associated with either drug when used individually.
\end{itemize}
These labels are based on established pharmacological theory~\cite{rang2011rang, ellison2002goodman}.
Drug pairs that do not fall into any of the above categories are considered to exhibit \textbf{no or unclear interaction}, indicating a lack of evidence or insufficient characterization of their pharmacodynamic relationship.

Figure~\ref{fig:data_pipeline} provides an overview of the MUDI dataset construction pipeline, illustrating both the data collection and semi-automated annotation phases along with their respective steps.


\subsection{Data Collection Phase}
\label{sec:datacollection}

\paragraph{\textbf{Eligibility Filtering:}}
We begin with a comprehensive list of drugs from DrugBank (version 5.1.12)~\cite{wishart2018drugbank}.
This list is refined using Hetionet version 1.0~\cite{himmelstein2017systematic}, which includes only drugs approved for human use and excludes experimental, toxic, or veterinary compounds.
In addition, Hetionet prioritizes drugs with clear therapeutic indications and biomedical connectivity, improving clinical relevance and downstream integration.

\paragraph{\textbf{Multimodal Collection:}}
For each eligible drug, we extract four data modalities from DrugBank.
Drugs missing any modality are excluded.
Four types of modalities are extracted or generated:
\begin{itemize}[leftmargin=2em]
    \item \textit{Textual modality}: Includes drug name and descriptions (summary, indications, pharmacodynamics, mechanism of action, and metabolism).
    Note that the drug-drug interaction fields in the metadata were excluded from the textual modality.
    
    \item \textit{Molecular structure graph}: SMILES (simplified molecular input line entry system) strings are processed with RDKit~\cite{landrum2013rdkit} into graphs where atoms are nodes and bonds are edges.
    
    \item \textit{Molecular structure image }: The processed SMILES were also used to generate standardized 2D chemical structure images rendered at a fixed resolution.
    
    \item \textit{Chemical formula}: Formulas are converted into normalized sequences with explicit atomic counts (e.g., `C$_{10}$H$_{13}$N$_5$O$_4$' becomes `\textit{Carbon 10 Hydrogen 13 Nitrogen 5 Oxygen 4}').
\end{itemize}

\begin{figure}[!t]
    \centering
    \includegraphics[width=\linewidth]{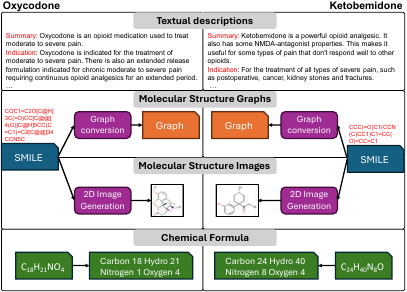}
    \caption{Multimodal examples.} 
    \Description{
       Multimodal representation of two drugs: Oxycodone and Ketobemidone.
    }
    \label{fig:example}
\end{figure}

Figure~\ref{fig:example} illustrates an example of the multimodal representations for two drugs, \textit{Oxycodone} and \textit{Ketobemidone}.
The result of this step is referred to as \textbf{Filtered DrugBank} and is retained for a later metadata integration step. 

\paragraph{\textbf{DDI Sentences Extraction:}}
As mentioned above, the drug-drug interaction fields present in the DrugBank metadata are excluded from the textual modality and handled separately.
In this step, we extract and separate this field into individual sentences that explicitly describe the interaction relationships between drug pairs.
These extracted sentences serve as the basis for downstream annotation.

\subsection{Semi-automated Annotation Phase}
\label{sec:data_annotation}
It is called the semi-automated annotation phase, as the process is primarily automated but involves human validation: linguistic annotators corrected masking errors, and medical experts curated the final labels.

\paragraph{\textbf{Drug Mention Alignment:}}  
The input to this step is a DDI sentence describing an interaction between a known pair of drugs.
Since DrugBank does not provide explicit tags linking drug names to their textual mentions within the sentence, we perform automatic alignment to identify the specific text spans referring to the two interacting drugs.
To achieve this, we apply flexible matching against all known synonyms of each drug. 

\paragraph{\textbf{Drug Masking:}}  
Once drug mentions are aligned, their corresponding text spans are replaced with placeholders \texttt{[DRUG1]} and \texttt{[DRUG2]}.
This masking abstracts away lexical differences across sentences and facilitates more consistent downstream classification. 
As a result, the original set of $244{,}921$ DDI descriptions was reduced to 287 distinct masked sentence templates.
However, due to limitations in the automated drug mention alignment step, some sentences contained incorrect masking -- such as partial drug names or missed detections when names overlapped (e.g., Promazine and Acepromazine).
These sentences were manually reviewed and corrected by three linguistics students, resulting in $241$ finalized sentence types.
This simple manual curation achieved near-perfect inter-annotator agreement.
See Appendix~\ref{appendix:masking_policy} for masking details.

\paragraph{\textbf{Pharmacodynamic Interaction Categorization:}}  
To assign pharmacodynamic labels, each masked sentence was automatically categorized into one of three predefined interaction types using a set of lexical rule-based heuristics.
These heuristics were derived from recurring linguistic templates and semantic patterns observed in the DrugBank descriptions.
For example, expressions indicating increased concentration or efficacy were mapped to \texttt{Synergism}, while patterns suggesting reduced activity were assigned to \texttt{Antagonism}.
Sentences that could not be confidently categorized were provisionally labeled as \texttt{New Effect}.
Full details of the heuristic rules and example templates are provided in Appendix~\ref{appendix:labeling_rules}.

\paragraph{\textbf{New Effect Validation:}} 
This step validates whether a \texttt{New Effect} truly refers to an adverse or therapeutic effect that is not present in either drug profiles.
We compare biomedical effect terms extracted from the DDI description sentences with the metadata of both drugs using stemming and synonym expansion techniques~\cite{porter1980algorithm, lesk1986automatic}.
If the effect term is absent from both drug profiles, the sentence is retained as a \texttt{New Effect}; otherwise, it is reclassified as \texttt{Synergism}.

\paragraph{\textbf{Expert Curation:}}  
To ensure biomedical accuracy, two physicians were involved in this expert curation phase.
They review all $241$ annotations and make corrections as needed based on the actual context and drug profiles.
In cases of disagreement, the two experts discussed to reach a final consensus.

\paragraph{\textbf{Metadata Integration:}} For each drug pair, we combine all collected modalities of two drugs, along with their interaction label, to form a complete data instance.
The resulting dataset is then split into training and test sets, with careful consideration to ensure that the test set contains novel examples for robust evaluation.

\subsection{Data Statistics and Analysis}
\label{sec:datastat}
The MUDI dataset contains $1{,}295$ unique drugs annotated with four modality types: structured text, SMILES sequences, molecular structure images, and graphs.
From these drugs, we generate $310{,}532$ labeled drug-drug interaction instances in three pharmacodynamic classes: \texttt{Synergism}, \texttt{Antagonism}, and \texttt{New Effect} (see Section~\ref{sec:data_annotation}).

\subsubsection{Label Distribution}
\label{subsec:label_distribution}

\begin{table}[!t]
\centering
\caption{Distribution of pharmacodynamic interaction labels in MUDI.}
\label{tab:label_distribution}
\resizebox{\linewidth}{!}{%
\begin{tabular}{lcccc}
    \hline
    \textbf{Label} & \textbf{Direction type} & \textbf{Train (\#)} & \textbf{Test (\#)} & \textbf{Total (\#)} \\
    \hline
    \multirow{2}{*}{Synergism} & Directed (uni) & 94,128 & 38,421 & 132,549 \\
                               & Directed (bi)  & 92,140 & 32,580 & 124,720 \\
    Antagonism                 & Directed (uni) & 27,320 & 11,914 & 39,234 \\
    New Effect                 & Undirected     & 7,527  & 6,502  & 14,029 \\
    \hline
    \multicolumn{2}{r}{\textbf{Total}}          & \textbf{221,115} & \textbf{89,417} & \textbf{310,532} \\
    \hline
\end{tabular}%
}
\end{table}

Table~\ref{tab:label_distribution} summarizes the label distribution across train and test sets.
The majority ($82.85\%$) of samples are \texttt{Synergism}, followed by \texttt{Antagonism} ($12.63\%$) and \texttt{New Effect} ($4.52\%$).
It reflects real-world pharmacological distribution, where synergistic effects are more frequently reported, while new effect interactions remain rare~\cite{harpaz2012novel}.
This natural imbalance encourages the design of robust classification strategies.

In addition to label variety, MUDI introduces \textbf{direction-aware annotations} -- a key novelty over prior datasets.
While \texttt{Antagonism} is strictly uni-directional and \texttt{New Effect} is undirected, some \texttt{Synergism} instances are bi-directional, allowing nuanced modeling of asymmetric and reciprocal effects.

\subsubsection{Drug Pair Coverage and Generalization}
\label{subsec:drug_generalization}

To assess the model's capacity for generalization, we explicitly construct a test set containing drug combinations that are underrepresented or entirely unseen during training.
Table~\ref{tab:pair_generalization} categorizes test drug pairs based on the presence of their constituent drugs in the training set.

\begin{table}[!t]
    \centering
    \caption{Generalization coverage of test drug pairs based on training set exposure.}
    \label{tab:pair_generalization}
    \begin{tabular}{lcc}
        \hline
        \textbf{Drug Pair Type} & \textbf{Count} & \textbf{Proportion} \\
        \hline
        Both drugs seen in training & 4,099 & 4.58 \% \\
        One drug seen in training & 78,822 & 88.15 \% \\
        Neither drug seen in training & 6,496 & 7.27 \% \\
        \hline
    \end{tabular}
\end{table}

The majority of test interactions ($88.15\%$) involve drug pairs where only one drug is familiar to the model, creating a semi-unseen generalization scenario.
Only $4.58\%$ of test pairs contain two seen drugs, while $7.27\%$ involve entirely unseen drugs, enabling rigorous evaluation of zero-shot prediction.
This careful partitioning strategy enables the assessment of generalization capabilities, reflecting practical challenges in biomedical applications where new drug combinations are continuously emerging \cite{zhang2020deep}.


\subsubsection{Modality-Level Data Profiling}
\label{subsec:modality_analysis}

Each drug entity in MUDI is represented through multiple modalities, each offering unique biological or structural information.
We analyze their properties to provide insights for modality-specific and fusion-based models.

\begin{figure}[!t]
    \centering
    \includegraphics[width=\linewidth]{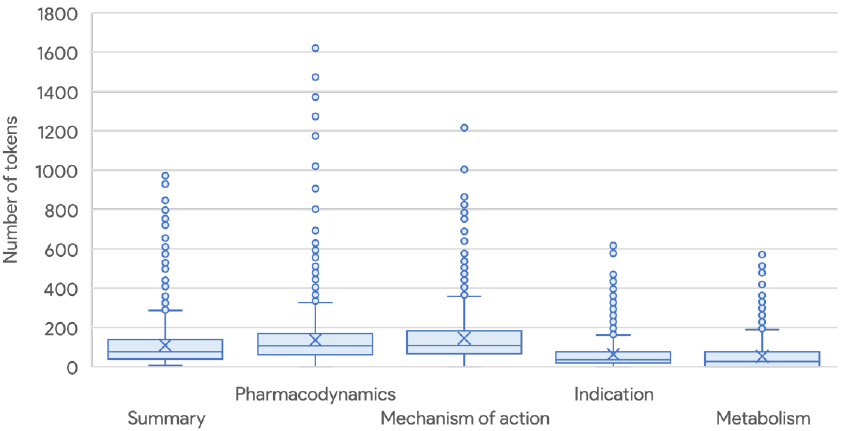}
    \caption{Token length distributions for five pharmacological text fields in MUDI.}
    \label{fig:text_length_distribution}
    \vspace{-.1cm}
\end{figure}

\paragraph{Pharmacological Text Descriptions}
Token length distributions across five fields are shown in Figure~\ref{fig:text_length_distribution}.
%
\textit{Pharmacodynamics}, \textit{Summary}, and \textit{Mechanism of Action} exhibit heavy-tailed lengths, often exceeding $1{,}000$ tokens.
In contrast, \textit{Indication} and \textit{Metabolism} are concise and consistent, often comprising under 100 tokens.
These variations call for adaptable text encoding strategies that can accommodate both long-form and short-form biomedical content.


\begin{table}[!t]
    \centering
    \caption{Statistics of molecular graphs derived from SMILES representations.}
    \label{tab:graph_statistics}
    \vspace{-.2cm}
    \begin{tabular}{lcc}
        \hline
        \textbf{Statistic} & \textbf{Mean} & \textbf{Std. Dev.} \\
        \hline
        Atoms per graph & 27.06 & 15.00 \\
        Bonds per graph & 27.79 & 16.25 \\
        Average node degree & 2.10 & 0.22 \\
        \hline
    \end{tabular}
    \vspace{-.3cm}
\end{table}

\paragraph{Molecular Graph Properties}
Table~\ref{tab:graph_statistics} reports statistics for SMILES-derived graphs.
%
The average molecular graph in MUDI contains approximately 27 atoms and 28 bonds, with a node degree of 2.1, reflecting typical chemical sparsity.
However, the relatively high standard deviations suggest notable diversity in graph sizes and connectivity patterns across the dataset.
These properties motivate the use of graph neural networks that are robust to varying graph sizes and capable of capturing both local connectivity and long-range interactions. 
Additionally, models must handle sparsity efficiently to avoid overfitting on smaller or simpler structures while preserving signals from more complex compounds.


\paragraph{Chemical Structure Images}
Each drug's SMILES string is also rendered into a 1000$\times$800 resolution image.
These standardized 2D representations support consistent training for vision-based models like CNNs and transformers.

\subsubsection{Comparison with Existing DDI Resources}
\label{subsec:dataset_comparison}

\begin{table*}
\caption{Comparison of our MUDI dataset with existing DDI-related datasets and resources.}
\label{tab:data_comparison}
\vspace{-.3cm}
\resizebox{\textwidth}{!}{%
\begin{tabular}{llcccccccccc}
\hline
\multicolumn{1}{c}{\textbf{Dataset}} & \multicolumn{1}{c}{\textbf{Scope}} & \textbf{\begin{tabular}[c]{@{}c@{}}Data\\ size\end{tabular}} & \textbf{Drugs} & \textbf{\begin{tabular}[c]{@{}c@{}}Number\\ of\\ categories$^\S$\end{tabular}} & \textbf{\begin{tabular}[c]{@{}c@{}}Multi-\\ modal\end{tabular}} & \textbf{\begin{tabular}[c]{@{}c@{}}Textual\\ description\end{tabular}} & \textbf{SMILES} & \textbf{\begin{tabular}[c]{@{}c@{}}Chemical\\ formula\end{tabular}} & \textbf{\begin{tabular}[c]{@{}c@{}}Molecular\\ structure\\ image\end{tabular}} & \textbf{\begin{tabular}[c]{@{}c@{}}Molecular\\ structure\\ graph\end{tabular}} & \textbf{\begin{tabular}[c]{@{}c@{}}Directed\\ relations\end{tabular}} \\ \hline
DrugBank~\cite{wishart2018drugbank} & Drug information & 1,420,072$^*$ & 16,581$^\dagger$ & -- & -- & \checkmark & \checkmark & \checkmark & (\checkmark) & -- & -- \\
BioSNAP~\cite{biosnapnets} & Drug interaction & 48,514 & 1,514 & -- & -- & -- & -- & -- & -- & -- & -- \\
LIDDI~\cite{banda2015provenance} & Side effect & 103,774 & 345 & -- & -- & -- & -- & -- & -- & -- & -- \\
TDC~\cite{huang2021therapeutics} & DDI description & 191,808 & 1,706 & -- & -- & -- & \checkmark & -- & -- & -- & -- \\
DDInter~\cite{xiong2022ddinter} & Mechanism and risk & 236,834 & 1,833 & 10 & \checkmark & \checkmark & \checkmark & \checkmark & (\checkmark) & -- & -- \\
HODDI~\cite{wang2025hoddi} & Side effect & 109,744$^\ddagger$ & 2,506 & 1 & -- & -- & -- & -- & -- & -- & -- \\
Mendeley DDI~\cite{yu2020data} & DDI description & 222,696 & 1,868 & 20 & -- & -- & -- & -- & -- & -- & -- \\
DDI-2013~\cite{bjorne2013uturku} & Semantic DDI & 5,021 & 1,913 & 4 & -- & \checkmark & -- & -- & -- & -- & -- \\
TWOSIDES~\cite{tatonetti2012data} & Side effect & 63,473 & 645 & 1 & -- & -- & \checkmark & -- & -- & -- & -- \\ \hline
Our MUDI & Pharmacodynamics & 310,532 & 1,295 & 3 & \checkmark & \checkmark & \checkmark & \checkmark & \checkmark & \checkmark & \checkmark \\ \hline
\multicolumn{12}{r}{\begin{tabular}[c]{@{}r@{}}\footnotesize\textit{$^*$: DDI-related sentence. $^\dagger$: Only 4,532 drugs are involved in drug-drug interactions. $^\ddagger$: Negative samples are counted in the total. $^\S$: Number of positive class labels; 1 implies binary classification.}\\ \footnotesize\textit{(\checkmark): Data exists but was not included in the provided download.}\end{tabular}}
\end{tabular}%
}
\vspace{-.3cm}
\end{table*}

Table~\ref{tab:data_comparison} compares MUDI with leading DDI datasets in terms of size, modality coverage, and semantic detail.
MUDI is the only dataset that integrates \textbf{rich multimodal features} per drug and uses \textbf{directed pharmacodynamic interaction labels}.
Unlike most existing datasets, which are limited to single-modality inputs or binary classification tasks, MUDI supports multimodal learning and fine-grained reasoning over clinically meaningful interaction categories.

This combination of scale, multimodality, and semantic depth positions MUDI as a unique benchmark for evaluating robust, multimodal approaches to pharmacodynamic DDI prediction.

\subsection{Licensing, Access, and Ethics}

\paragraph{Open Release and Licensing}
To support transparency and reproducibility, MUDI is released under the Creative Commons Attribution-NonCommercial 4.0 License (CC BY-NC 4.0), which allows sharing and adaptation with attribution, but prohibits commercial use.
The dataset, preprocessing scripts, and baseline models are available on \href{https://zenodo.org/records/15544551}{Zenodo}\footnote{\url{https://zenodo.org/records/15544551}} and \href{https://github.com/hoangbros03/MUDI}{GitHub}\footnote{\url{https://github.com/hoangbros03/MUDI}}, along with documentation and full download instructions (Appendix~\ref{appendix:access}).

\vspace{-.1cm}
\paragraph{Ethical Considerations}
MUDI is built from publicly available biomedical sources (specifically, DrugBank~\cite{wishart2018drugbank}) and contains no human subject data or identifiable information, thus requiring no ethical review.
We encourage responsible use aligned with licensing terms and recommend expert consultation for clinical applications.
Responsible usage guidelines are provided in Appendix~\ref{appendix:ethical}.

\section{Baseline Experiments}
\label{sec:baselines}

\subsection{Baseline Models}
To benchmark MUDI and support future research, we implement a suite of baseline models covering both single-modality and multimodal fusion strategies.
Each model takes a pair of drugs as input and predicts one of three interaction types.
To train models effectively, we apply negative sampling by including 200,000 additional \textit{No Interaction} pairs as negative examples.
Implementation details and equations are provided in Appendix~\ref{appendix:baseline}.

\vspace{-.1cm}
\paragraph{Single-Modality Baselines}
We train six modality-specific models: (i) \texttt{name}, (ii) \texttt{description} (merged from summary, pharmacodynamics, mechanism, metabolism, and indication), (iii) \texttt{formula}, (iv) \texttt{SMILES} (text), (v) molecular structure \texttt{graph}, and (vi) \texttt{image}.
%
For \textbf{text-based} inputs (\texttt{name}, \texttt{description}, \texttt{SMILES}, \texttt{formula}), we fine-tune BioMedBERT~\cite{gu2021domain}, using the final \texttt{[CLS]} token as the drug embedding.
\textbf{Graph} inputs are processed using a graph convolutional network~\cite{kipf2017semi} with max pooling.
\texttt{\textbf{Image}} inputs are encoded using a Vision Transformer~\cite{dosovitskiy2021an}.
%
Each single-modality model concatenates two drug embeddings and feeds them to a shallow classifier.

\vspace{-.1cm}
\paragraph{Late Fusion Baseline}
We construct a late fusion model that aggregates predictions from six single-modality classifiers.
The final prediction is computed via majority voting across the six classifiers, with tie-breaking handled in a fixed modality priority order.
This baseline reflects ensembling without joint feature modeling.

\paragraph{Intermediate Fusion Baseline}
We construct an intermediate fusion baseline by concatenating the modality-specific embeddings into an interleaved sequence.
The resulting fused vector is passed through a two-layer MLP classifier.
This captures cross-modal interactions while preserving pairwise structure.


\subsection{Experimental Setup}
\label{sec:evaluation}


\paragraph{Evaluation Metrics.}
We report Precision, Recall, and F1 scores for each of the three positive classes, along with Micro and Macro averages.
Negative examples (\textit{No Interaction}) are included during training but excluded from positive-class metric computation, following common biomedical evaluation practice~\cite{peng2018extracting}.

\paragraph{Evaluation Settings.}
We evaluate under two settings:
(i)~\textbf{direction-aware} matching, where (\texttt{DRUG1}, \texttt{DRUG2}) $\neq$ (\texttt{DRUG2}, \texttt{DRUG1}),
and (ii)~\textbf{direction-agnostic} matching, where order is ignored.
Detailed evaluation scripts and protocols are provided in Appendix~\ref{appendix:evaluation}.

\paragraph{Implementation Environment and Hyper-parameters.}
All models are implemented in \texttt{PyTorch 3.10} and trained on NVIDIA T4 GPUs.
The training environment, model-specific hyper-parameters, and hardware details are fully documented in Appendix~\ref{appendix:environment_hyperparams}.

\subsection{Results}

\begin{table}[!t]
\caption{Multimodal baseline results on MUDI dataset.}
\label{tab:result_mudi}
\vspace{-.2cm}
\resizebox{\linewidth}{!}{%
\begin{tabular}{clccc}
\hline
\multicolumn{2}{c}{} & \textbf{Precision} & \textbf{Recall} & \textbf{F1} \\ \hline
\multicolumn{5}{c}{\textbf{Intermediate Fusion}} \\ \hline
\multirow{5}{*}{\begin{tabular}[c]{@{}c@{}}Direction-aware \\ matching\end{tabular}} & Synergism & 54.79 & 54.49 & 54.64 \\
 & Antagonism & 55.39 & 47.88 & 51.36 \\
 & New Effect & 61.68 & 17.97 & 27.84 \\
 & Micro-Averaged & 55.04 & 50.61 & 52.74 \\
 & Macro-Averaged & 57.29 & 40.11 & 47.18 \\ \hline
\multirow{5}{*}{\begin{tabular}[c]{@{}c@{}}Direction-agnostic \\ matching\end{tabular}} & Synergism & 78.98 & 62.47 & 69.76 \\
 & Antagonism & 71.61 & 45.32 & 55.51 \\
 & New Effect & 78.65 & 35.59 & 49.00 \\
 & Micro-Averaged & 77.91 & 58.29 & 66.69 \\
 & Macro-Averaged & 76.41 & 47.79 & 58.80 \\ \hline
\multicolumn{5}{c}{\textbf{Late Fusion}} \\ \hline
\multirow{5}{*}{\begin{tabular}[c]{@{}c@{}}Direction-aware \\ matching\end{tabular}} & Synergism & 38.32 & 70.82 & 49.73 \\
 & Antagonism & 48.12 & 36.12 & 41.26 \\
 & New Effect & 62.79 & 9.06 & 15.84 \\
 & Micro-Averaged & 39.25 & 60.32 & 47.56 \\
 & Macro-Averaged & 49.74 & 38.67 & 43.51 \\ \hline
\multirow{5}{*}{\begin{tabular}[c]{@{}c@{}}Direction-agnostic \\ matching\end{tabular}} & Synergism & 62.14 & 76.44 & 68.55 \\
 & Antagonism & 65.86 & 36.12 & 46.65 \\
 & New Effect & 76.97 & 17.99 & 29.16 \\
 & Micro-Averaged & 62.62 & 66.85 & 64.67 \\
 & Macro-Averaged & 68.32 & 43.52 & 53.17 \\  \hline
\end{tabular}%
}
\vspace{-.4cm}
\end{table}

Table~\ref{tab:result_mudi} presents the performance of our multimodal baselines under two evaluation settings: direction-aware and direction-agnostic matching.
Across all metrics, the results confirm the inherent difficulty of MUDI, particularly in predicting less frequent labels such as \texttt{New Effect}.
In the direction-aware setting, F1 scores are generally lower, with \texttt{New Effect} reaching only 27.84\% under intermediate fusion and 15.84\% under late fusion.
Notably, the exclusion of directionality during evaluation substantially improves performance.
For example, the micro-averaged F1 of intermediate fusion rises from 52.74\% (direction-aware) to 66.69\% (direction-agnostic).
This underscores the challenge of modeling subtle interactions and label imbalance.


Among the fusion strategies, intermediate fusion consistently outperforms late fusion across both settings. Its superior performance (up to 69.76\% F1 for \texttt{Synergism} and 66.69\% overall micro-F1) demonstrates the advantage of integrating multimodal features before final decision layers.
In contrast, late fusion appears less effective, likely due to the weak predictive capacity of individual unimodal branches in isolation.


To assess the contribution of each modality, we report single-modality results in Appendix~\ref{appendix:singlemodal}, Table~\ref{tab:single-modal-result}.
Graph-based features perform best, achieving 65.44\% micro-F1 under direction-agnostic matching.
Textual fields (e.g., name and description) and images also show competitive results.
However, SMILES and chemical formula underperform, indicating the need for more expressive encoders.

Together, these findings highlight (1) the need for robust multimodal integration to capture complementary biomedical signals, and (2) the utility of MUDI in benchmarking model generalization and cross-modal reasoning under realistic constraints.

\section{Discussion}

\paragraph{Potential Applications}
The MUDI dataset supports several promising research directions within multimodal biomedical machine learning.
First, it serves as a robust resource for developing models aimed at \textbf{early detection of pharmacodynamic interactions}, which can help researchers better understand complex drug behaviors.
MUDI can also be leveraged in developing \textbf{DDI-aware recommendation systems} to assist researchers and pharmacists in preclinical compound screening and pharmaceutical development, as well as preliminary \textbf{interaction risk screening} for early-stage drug discovery.
Additionally, MUDI enables the creation of \textbf{educational tools} for training healthcare professionals and pharmacologists on multimodal biomedical data interpretation.
Its inclusion of unseen drug pairs further supports research into \textbf{generalizable prediction methods} for novel or investigational compounds, thereby promoting innovation in zero-shot biomedical learning.
Nevertheless, given the dataset's limitations in clinical validation and potential label noise, models trained on MUDI should not be directly applied to clinical or regulatory decisions (see Appendix~\ref{appendix:ethical}).
Therefore, appropriate usage involves foundational research, educational training, or benchmarking within controlled experimental conditions, rather than immediate clinical implementation.

\paragraph{Insights}
The baseline results provide valuable insights into multimodal modeling on MUDI.
First, intermediate fusion, which integrates multimodal information earlier within the model, consistently outperforms late fusion strategies across all evaluation settings, underscoring the importance of early interactions between modalities.
Second, performance significantly improves when directional constraints are relaxed during evaluation, particularly for the \textit{Synergism} class, indicating that directional ambiguity is an important factor affecting model predictions.
Lastly, the single-modality analysis (Appendix~\ref{appendix:singlemodal}) highlights the strong predictive capability of molecular graph representations relative to other modalities, suggesting the particular utility of structural information in predicting pharmacodynamic interactions.

\paragraph{Limitations}
Despite careful curation, MUDI presents some limitations.
A key constraint is the absence of \textbf{gold-standard validation} from clinical trials or wet-lab experiments.
Interaction labels are derived from structured drug information and, although carefully annotated, may not fully reflect the complexity of \textbf{real-world clinical practice}. 
Additionally, reliance on textual descriptions introduces potential label noise, as \textbf{linguistic ambiguities or inconsistencies} in drug metadata can lead to occasional mislabeling~\cite{bodenreider2004unified}.

\paragraph{Future Work}
Several directions exist to further extend MUDI's utility and scientific impact.
Future work could expand the dataset with separate \textbf{pharmacokinetic} labels (e.g., absorption, metabolism) to support \textbf{multi-label} interaction modeling and disentangle the dynamics between pharmacokinetics and pharmacodynamics.
Another direction is integrating \textbf{biomedical knowledge graphs}~\cite{zhou2020knowledge} to enrich drug representations, support structured reasoning, and enhance both performance and explainability.

\section{Conclusion}
In this paper, we (1) introduce MUDI, a large-scale, multimodal biomedical dataset for predicting pharmacodynamic interactions between drug-drug interactions (DDIs) and (2) benchmark competitive machine learning to predict DDI using MUDI. MUDI addresses key gaps in existing resources by integrating pharmacological text, molecular graphs, and chemical structure images, providing a comprehensive multimodal data source for investigating complex interactions between medications.
Additionally, MUDI includes previously unseen drugs in the test set, thereby supporting robust evaluation of machine learning models' generalization capabilities beyond known drug pairs.

The strengths of MUDI lie in its scale, multimodal richness, real-world clinical relevance, benchmarking results, and (free) availability.
By covering diverse modalities and focusing specifically on pharmacodynamic effects, MUDI enables the development of potentially more accurate, generalizable, and targeted DDI prediction models. 
Further, MUDI's structured annotation guidelines, open-access licensing, and reproducibility protocols enhance its utility for research, clinics, and potentially, drug discovery.

Exploring MUDI, future work may gain further insights into complex interactions between drugs. Future work may also use MUDI to improve medication assignment when treating comorbidities, and, potentially, based on selected multimodal features predictive DDI effects, improve drug development.



\bibliographystyle{ACM-Reference-Format}
\bibliography{refs}

\clearpage

\appendix

\section{Annotation Guidelines}
\label{appendix:annotation}

This section describes the manual annotation protocol used to label drug-drug interaction (DDI) pairs in the MUDI dataset. Our objective is to create a standardized, clinically meaningful categorization of pharmacodynamic interactions into three classes: \texttt{Synergism}, \texttt{Antagonism}, and \texttt{New Effect}.

\subsection{Drug Name Masking Policy}
\label{appendix:masking_policy}

To ensure consistent pattern recognition and reduce annotator bias toward specific drug identities, all drug mentions in the original DrugBank interaction descriptions are replaced with abstract placeholders before annotation.
This masking step is essential for allowing models and human annotators to focus on the nature of the pharmacodynamic interaction rather than the specific lexical forms of drug names.

\paragraph{Standardized Placeholders.}  
The two interacting drugs in each sentence are masked using \texttt{[DRUG1]} and \texttt{[DRUG2]}.
Any additional drug names appearing in the same sentence are replaced with \texttt{[DRUGOTHER]}.
This abstraction is applied to both brand names and generic names, as well as chemical synonyms.

\paragraph{Drug Mention Alignment.}  
Since DrugBank does not provide token-level alignment between drug entities and their textual positions, we employ a flexible string-matching algorithm to detect mentions of each drug and its synonyms.
This process uses:
\begin{itemize}[leftmargin=2em]
    \item Canonical names and aliases from DrugBank metadata.
    \item Case-insensitive matching.
    \item Partial overlap resolution (to disambiguate e.g., ``Promazine'' vs. ``Acepromazine'').
\end{itemize}

\paragraph{Manual Review and Correction.}  
In some cases, automatic matching resulted in incomplete or incorrect spans -- especially when drugs shared lexical substrings or when spacing/punctuation was irregular.
To address this, a team of three linguistics students manually reviewed and corrected all unique masked sentence templates.
The original corpus of $244{,}921$ raw DDI descriptions was thereby reduced to 287 distinct masked templates, from which 241 final sentence types were retained after quality filtering.
Inter-annotator agreement during this manual review phase was near-perfect, with disagreements resolved by discussion and unanimous consensus.

\paragraph{Impact on Annotation Quality.}  
This masking strategy eliminates the risk of model or annotator bias due to prior familiarity with drug names or brand-specific expectations.
By enforcing a uniform abstracted representation across the dataset, we enable more consistent labeling of pharmacodynamic effects and allow models to generalize beyond known drug pairs.

\subsection{Labeling Rules}
\label{appendix:labeling_rules}
Interaction descriptions in MUDI are categorized into one of three pharmacodynamic classes based on a set of carefully constructed lexical heuristics.
These heuristics are grounded in recurring sentence structures observed in DrugBank and designed to reflect pharmacological theory~\cite{rang2011rang, ellison2002goodman}.


\paragraph{Synergism.}  
Labeled when \texttt{[DRUG1]} enhances the pharmacological effect, bioavailability, or systemic concentration of \texttt{[DRUG2]}. Typical patterns include increased absorption, inhibited excretion, or elevated therapeutic efficacy.

\vspace{2mm}
\noindent\textbf{Rule Templates for \texttt{Synergism}:}
\begin{itemize}[leftmargin=2em]
    \item \texttt{[DRUG1]} can cause an increase in the absorption of \texttt{[DRUG2]} resulting in an increased serum concentration and potentially a worsening of adverse effects.
    \item \texttt{[DRUG1]} may decrease the excretion rate of \texttt{[DRUG2]} which could result in a higher serum level.
    \item \texttt{[DRUG1]} may increase the \texttt{[activities names]} activities of \texttt{[DRUG2]}.
    \item The bioavailability of \texttt{[DRUG1]} can be increased when combined with \texttt{[DRUG2]}.
    \item The excretion of \texttt{[DRUG1]} can be decreased when combined with \texttt{[DRUG2]}.
    \item The metabolism of \texttt{[DRUG1]} can be decreased when combined with \texttt{[DRUG2]}.
    \item The protein binding of \texttt{[DRUG1]} can be decreased when combined with \texttt{[DRUG2]}.
    \item The serum concentration of \texttt{[DRUG1]} can be increased when it is combined with \texttt{[DRUG2]}.
    \item The serum concentration of \texttt{[metabolite name]}, an active metabolite of \texttt{[DRUG1]}, can be increased when used in combination with \texttt{[DRUG2]}.
    \item The therapeutic efficacy of \texttt{[DRUG1]} can be increased when used in combination with \texttt{[DRUG2]}.
\end{itemize}


\paragraph{Antagonism.}  
An interaction is labeled as \texttt{Antagonism} if the description suggests that \texttt{[DRUG1]} inhibits, reduces, or interferes with the pharmacodynamic action of \texttt{[DRUG2]}, including diminished absorption, faster metabolism, or decreased efficacy.

\vspace{2mm}
\noindent\textbf{Rule Templates for \texttt{Antagonism}:}
\begin{itemize}[leftmargin=2em]
    \item \texttt{[DRUG1]} can cause a decrease in the absorption of \texttt{[DRUG2]} resulting in a reduced serum concentration and potentially a decrease in efficacy.
    \item \texttt{[DRUG1]} may decrease effectiveness of \texttt{[DRUG2]} as a diagnostic agent.
    \item \texttt{[DRUG1]} may decrease the \texttt{[activities names]} activities of \texttt{[DRUG2]}.
    \item \texttt{[DRUG1]} may increase the excretion rate of \texttt{[DRUG2]} which could result in a lower serum level and potentially a reduction in efficacy.
    \item The absorption of \texttt{[DRUG1]} can be decreased when combined with \texttt{[DRUG2]}.
    \item The bioavailability of \texttt{[DRUG1]} can be decreased when combined with \texttt{[DRUG2]}.
    \item The excretion of \texttt{[DRUG1]} can be increased when combined with \texttt{[DRUG2]}.
    \item The metabolism of \texttt{[DRUG1]} can be increased when combined with \texttt{[DRUG2]}.
    \item The risk or severity of \texttt{[adverse effects]} can be decreased when \texttt{[DRUG1]} is combined with \texttt{[DRUG2]}.
    \item The serum concentration of \texttt{[DRUG1]} can be decreased when it is combined with \texttt{[DRUG2]}.
    \item The serum concentration of \texttt{[metabolite name]}, an active metabolite of \texttt{[DRUG1]}, can be decreased when used in combination with \texttt{[DRUG2]}.
    \item The therapeutic efficacy of \texttt{[DRUG1]} can be decreased when used in combination with \texttt{[DRUG2]}.
\end{itemize}


\paragraph{New Effect.}
An interaction is labeled as \texttt{New Effect} when the interaction leads to a novel adverse effect not known to be associated with either \texttt{[DRUG1]} or \texttt{[DRUG2]} independently. 
Initially, an interaction assigned the \texttt{New Effect} label when the sentence contains biomedical event terms (typically adverse effects) that cannot be clearly attributed to either \texttt{Synergism} or \texttt{Antagonism} patterns. These are typically masked sentences that do not indicate enhancement or suppression but instead describe the emergence of a distinct pharmacological effect.

To validate whether the reported effect is indeed novel to the drug combination, we perform a comparison between the extracted biomedical term and the known side effect profiles of both drugs. This process includes:
\begin{enumerate}
    \item Automatically identifying the biomedical effect term in the masked sentence.
    \item Matching the term against the adverse event metadata of each drug using stemming and synonym expansion~\cite{porter1980algorithm, lesk1986automatic}.
    \item Assigning the \texttt{New Effect} label only if the effect is absent from both drug profiles.
    \item Reassigning the sentence to \texttt{Synergism} if the effect is already associated with one of the drugs.
\end{enumerate}

All candidate \texttt{New Effect} annotations are subsequently reviewed by domain experts to ensure biomedical validity. This step ensures that no pharmacologically implausible or redundant labels remain in the final dataset.

\paragraph{No or Unclear Interaction.}
Interaction descriptions that do not exhibit clear pharmacodynamic effects -- such as therapeutic enhancement, attenuation, or novel adverse outcomes -- are not assigned a positive label.
This includes cases with vague, incomplete, or purely pharmacokinetic information lacking clinical consequence.
In line with established pharmacological theory~\cite{rang2011rang, ellison2002goodman}, such drug pairs are considered to exhibit \textit{no or unclear interaction}, indicating insufficient evidence to support classification into one of the defined pharmacodynamic categories.



    



\subsection{Annotation Quality Control}

To ensure the biomedical validity and consistency of interaction labels in MUDI, we implemented a two-phase expert curation protocol.
After automated annotation using lexical heuristics (Section~\ref{sec:data_annotation}), each of the 241 distinct masked interaction templates was reviewed by two physicians with domain expertise in clinical pharmacology and drug safety.

Each expert independently examined the interaction context, checked consistency with known drug properties, and validated the correctness of the assigned label with respect to both pharmacodynamic semantics and biomedical relevance.
In cases where the experts initially disagreed, they conducted a focused discussion to reach a consensus.
This adjudication phase ensured the removal of residual annotation noise introduced by the automated pipeline, particularly in borderline or semantically ambiguous cases.

This dual-review and consensus-based process ensures that MUDI maintains clinically credible labels and supports reliable downstream model development.

\section{Dataset Access and Organization}
\label{appendix:access}

This appendix provides practical information for obtaining and using the MUDI dataset, including access instructions, licensing terms, and directory structure.

\subsection{Access and Licensing}

The MUDI dataset is derived from DrugBank, a publicly accessible biomedical database. According to DrugBank's data usage policy\footnote{\url{https://go.drugbank.com/legal}}, academic and non-commercial use of DrugBank content is permitted under its custom license for research purposes.
In full compliance with this policy, MUDI builds on openly accessible DrugBank fields and restricts redistribution to non-commercial academic use only.

We release the MUDI dataset under the Creative Commons Attribution-NonCommercial 4.0 International License (CC BY-NC 4.0)\footnote{\url{https://creativecommons.org/licenses/by-nc/4.0/}}.
This license enables researchers to copy, distribute, and adapt the dataset for academic purposes, provided that:
\begin{itemize}[leftmargin=2em]
    \item Proper credit is given to both the MUDI project and the original DrugBank resource.
    \item Any derived works or models are clearly marked as adaptations.
    \item No commercial use is made without explicit written permission.
\end{itemize}

Users may download the dataset, documentation, preprocessing scripts, and baseline code from the following links:
\begin{itemize}[leftmargin=2em]
    \item \textbf{Zenodo}: \url{https://zenodo.org/records/15544551} -- the dataset archive with DOI for stable citation.
    \item \textbf{GitHub}: \url{https://github.com/hoangbros03/MUDI} -- the codebase repository for preprocessing, baseline models, and future updates.
\end{itemize}
A permanent DOI link\footnote{\url{https://doi.org/10.5281/zenodo.15544551}} is assigned via Zenodo to ensure stable referencing and citation.

By downloading and using MUDI, users agree to comply with the license terms and responsible usage guidelines outlined in Appendix~\ref{appendix:ethical}.

\subsection{Dataset Structure}

The MUDI dataset is organized into a clear and modular directory layout to facilitate ease of use, reproducibility, and multimodal experimentation:

\vspace{.2cm}
\includegraphics[width=.53\linewidth]{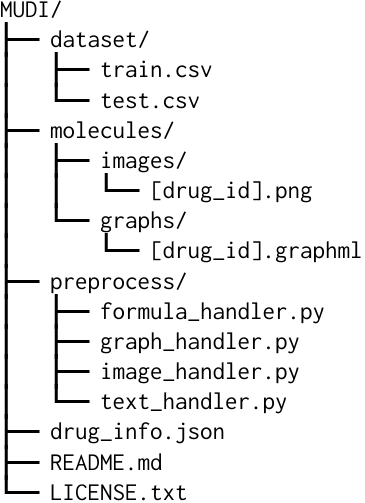}

\paragraph{\textbf{\texttt{dataset/train.csv} and \texttt{test.csv}}}
Each row in these CSV files corresponds to a labeled drug-drug interaction (DDI) instance and contains the following fields:
\begin{itemize}[leftmargin=2em]
    \item \texttt{DRUG1}: Unique identifier for the first drug.
    \item \texttt{Interaction}: One of the pharmacodynamic classes (\texttt{Synergism}, \texttt{Antagonism}, or \texttt{New Effect}).
    \item \texttt{DRUG2}: Unique identifier for the second drug.
\end{itemize}
All drug identifiers are keys in \texttt{drug\_info.json} for retrieving textual, structural, and molecular representations.

\paragraph{\textbf{\texttt{molecules/images/}}}
Each PNG image represents the 2D chemical structure of a drug generated from its SMILES string using RDKit. All images are:
\begin{itemize}[leftmargin=2em]
    \item Named as \texttt{[drug\_id].png}.
    \item Stored at a standardized resolution of 1000$\times$800 pixels.
    \item Ready for direct use with image encoders such as Vision Transformer.
\end{itemize}

\paragraph{\textbf{\texttt{molecules/graphs/}}}
Each file is a GraphML-encoded molecular graph where:
\begin{itemize}[leftmargin=2em]
    \item Nodes represent atoms.
    \item Edges represent bonds (e.g., single, double, aromatic).
\end{itemize}
The files follow the standard GraphML format, with recommended compatibility via the \texttt{NetworkX} Python library.

\paragraph{\textbf{\texttt{drug\_info.json}}}
This JSON file consolidates metadata for each drug. Each entry contains:
\begin{itemize}[leftmargin=2em]
    \item \texttt{name}: Human-readable drug name, e.g., ``Amitriptyline''.
    
    \item \texttt{description}: A dictionary containing pharmacological text fields used for textual modeling:
    \begin{itemize}[leftmargin=1em]
        \item \texttt{summary}: A concise overview of the drug's identity, primary use, and general characteristics.
        \item \texttt{indication}: Approved medical conditions or diseases that the drug is prescribed to treat.
        \item \texttt{metabolism}: Description of the drug's metabolic pathway, including hepatic enzymes involved (e.g., CYP450 family).
        \item \texttt{pharmacodynamics}: Explanation of the biological effects, mechanism of drug action, and dose-response relationships.
        \item \texttt{moa} (mechanism of action): Detailed molecular-level explanation of how the drug achieves its intended effect, such as receptor binding or enzyme inhibition.
    \end{itemize}
    
    \item \texttt{formula}: The molecular formula representing the elemental composition of the drug (e.g., \texttt{C20H25N3O}).
    
    \item \texttt{smiles}: The canonical SMILES (Simplified Molecular Input Line Entry System) string encoding the molecular structure in a compact, text-based format.
\end{itemize}
These fields are used to build modality-specific representations for textual, formula-based, and structural modeling.

\subsection{File Standards and Preprocessing Code}

\paragraph{File Formats.}
\begin{itemize}[leftmargin=2em]
    \item All interaction data is UTF-8 encoded CSV.
    \item Molecular graphs follow the GraphML standard.
    \item Chemical structure diagrams are stored as PNG images.
    \item Metadata is provided as structured JSON.
\end{itemize}

\paragraph{Preprocessing Scripts.}
The \texttt{preprocess/} directory includes modular Python scripts for converting raw inputs into model-ready formats:
\begin{itemize}[leftmargin=2em]
    \item \texttt{text\_handler.py}: Create a JSON object holding textual information, including name, description, SMILES, and formula.
    \item \texttt{formula\_handler.py}: Improve the representation of molecule elements within chemical formulas before they are passed to the text handler.
    \item \texttt{image\_handler.py}: Load the images from the dataset and convert them into tensors.
    \item \texttt{graph\_handler.py}: Load and create graph objects.
\end{itemize}
Key dependencies include \texttt{RDKit}, \texttt{networkx}, and \texttt{transformers}.

\subsection{Getting Started}

To quickly begin using MUDI:
\begin{itemize}[leftmargin=2em]
    \item Refer to \texttt{README.md} for installation, tutorials, and citation information.
    \item Use the \texttt{drug\_info.json} file to retrieve all relevant metadata for each drug.
    \item Apply the preprocessing scripts to regenerate modality-specific features as needed.
    \item Evaluate models using the provided train/test splits and compute metrics such as precision, recall, micro-F1, and macro-F1.
\end{itemize}
The provided setup is fully reproducible and extensible for future multimodal biomedical research.

\section{Responsible Usage and Ethical Guidelines}
\label{appendix:ethical}

This appendix details the ethical foundations, responsible usage requirements, and recommended best practices for working with the MUDI dataset.
These principles aim to promote transparency, safety, and compliance in biomedical AI research.

\subsection{Data Provenance and Privacy}

The MUDI dataset is built entirely from non-sensitive, publicly accessible data obtained from the DrugBank database~\cite{wishart2018drugbank}, a reputable biomedical resource.
All data sources are governed by DrugBank's academic use policy\footnote{\url{https://go.drugbank.com/legal}}, which permits reuse for non-commercial research.

Importantly, MUDI does not contain any protected health information (PHI), patient-level records, or personally identifiable information (PII).
No data originates from clinical trials, electronic medical records, or real-world hospital systems.
As such, the dataset does not fall under the scope of human subjects research and does not require ethical approval from an institutional review board (IRB).
It is also exempt from compliance obligations under HIPAA, GDPR, or related data privacy regulations.

\subsection{Intended Use}

MUDI is intended exclusively for academic research, education, and non-commercial purposes.
Acceptable use cases include, but are not limited to:
\begin{itemize}[leftmargin=2em]
    \item Development, benchmarking, and publication of multimodal learning algorithms for biomedical knowledge discovery.
    \item Research in drug-drug interaction (DDI) prediction, representation learning, cross-modal retrieval, and zero-shot biomedical reasoning.
    \item Classroom use in university-level courses or technical workshops on machine learning, drug discovery, or bioinformatics.
\end{itemize}

Any commercial use of the dataset is prohibited under the terms of the CC BY-NC 4.0 license without explicit written permission from the authors.

\subsection{Known Limitations and Usage Caveats}

Despite thorough curation and validation, MUDI remains a research-focused dataset and carries certain limitations:
\begin{itemize}[leftmargin=2em]
    \item \textbf{No Clinical Validation:} Pharmacodynamic interaction labels are generated through lexical rules and expert curation, but not independently verified in wet-lab or clinical settings. The dataset is not intended for clinical use or decision support.
    
    \item \textbf{Potential Label Ambiguity:} The source descriptions from DrugBank are natural language statements, which may contain implicit or ambiguous interaction signals. While the annotation pipeline includes validation steps, some residual label noise is inevitable.
    
    \item \textbf{Pharmacodynamic Scope Only:} MUDI exclusively targets pharmacodynamic interactions. It does not cover pharmacokinetic DDIs such as those involving absorption, distribution, metabolism, or excretion (ADME) pathways.
    
    \item \textbf{Bias Toward Common Drugs:} Interaction labels are inherently more complete for well-studied drugs. This may bias model performance toward drugs with richer metadata and documented histories.
\end{itemize}
Researchers should exercise caution when interpreting results for clinical decision support or downstream biomedical applications.

\subsection{Responsible Research Practices}

We encourage users of MUDI to adopt the following practices to uphold ethical standards and maximize the scientific value of their work:
\begin{itemize}[leftmargin=2em]
    \item Clearly cite the MUDI dataset and its associated publication in all derivative research.
    \item Disclose all modeling assumptions, training data subsets, and evaluation procedures to support reproducibility.
    \item Publicly release code and model checkpoints when possible, subject to the same licensing terms.
    \item Transparently communicate dataset limitations, especially when proposing real-world or clinical applications.
    \item Avoid deploying or advertising models trained on MUDI for direct use in patient care without formal clinical validation and regulatory approval.
\end{itemize}

\section{Baseline Model Configurations}
\label{appendix:baseline}

This appendix provides detailed descriptions of the baseline models used to benchmark the MUDI dataset, including architecture choices, modality-specific preprocessing, and mathematical formulations.

\subsection{Single-Modality Baselines}

\subsubsection{Text-only Baseline (BioMedBERT)}

The text-only model uses BioMedBERT~\cite{gu2021domain} to encode concatenated pharmacological fields: summary, indication, mechanism of action, pharmacodynamics, and metabolism.
The fields are joined into a single input sequence, separated by special tokens.

Given an input sequence $\mathbf{x}^{\text{text}}$, the model computes hidden representations $\mathbf{h}_i$ for each token:
\begin{equation*}
\mathbf{h}_i = \mathcal{E}_{\text{BioMedBERT}}(\mathbf{x}^{\text{text}})_i,
\end{equation*}
where $i$ indexes the tokens.

We extract the embedding corresponding to the \texttt{[CLS]} token, $\mathbf{h}_{\texttt{[CLS]}}$, and apply a linear classification layer:
\begin{equation*}
\hat{\mathbf{y}}_{\text{text}} = \text{softmax}(\mathbf{h}_{\text{[CLS]}} \mathbf{W}_t + \mathbf{b}_t),
\end{equation*}
where $\mathbf{W}_t \in \mathbb{R}^{d \times C}$ and $\mathbf{b}_t \in \mathbb{R}^{C}$ are learnable parameters, $d$ is the embedding dimension, $C$ is the number of output classes, and $\hat{\mathbf{y}}_{\text{text}} \in \mathbb{R}^C$ is the predicted class distribution.

\subsubsection{Graph-only Baseline (GCN)}

The graph-only model uses a two-layer Graph Convolutional Network (GCN)~\cite{kipf2017semi} to process the molecular structure graphs generated from SMILES strings.

Each molecule graph is represented as an adjacency matrix $\mathbf{A}$ and a feature matrix $\mathbf{X}$ containing atom features.
The GCN updates node features as:
\begin{align*}
\mathbf{H}^{(1)} &= \sigma\left(\tilde{\mathbf{D}}^{-\frac{1}{2}} \tilde{\mathbf{A}} \tilde{\mathbf{D}}^{-\frac{1}{2}} \mathbf{X} \mathbf{W}^{(0)}\right),
\\
\mathbf{H}^{(2)} &= \sigma\left(\tilde{\mathbf{D}}^{-\frac{1}{2}} \tilde{\mathbf{A}} \tilde{\mathbf{D}}^{-\frac{1}{2}} \mathbf{H}^{(1)} \mathbf{W}^{(1)}\right),
\end{align*}
where $\tilde{\mathbf{A}} = \mathbf{A} + \mathbf{I}$ is the adjacency matrix with self-loops added, $\tilde{\mathbf{D}}$ is the corresponding degree matrix, and $\sigma$ denotes the ReLU activation function. The learnable parameters $\mathbf{W}^{(0)} \in \mathbb{R}^{d_{\text{in}} \times d_{\text{hidden}}}$ and $\mathbf{W}^{(1)} \in \mathbb{R}^{d_{\text{hidden}} \times d_{\text{out}}}$ are weight matrices for the first and second GCN layers, respectively, where $d_{\text{in}}$ is the input node feature dimension, $d_{\text{hidden}}$ is the hidden dimension, and $d_{\text{out}}$ is the output node feature dimension.

After the second GCN layer, we obtain node-level embeddings $\mathbf{H}^{(2)} = [\mathbf{h}_1^{(2)}, \mathbf{h}_2^{(2)}, \ldots, \mathbf{h}_n^{(2)}]$, where $\mathbf{h}_i^{(2)} \in \mathbb{R}^d$ is the feature vector of the $i$-th node and $n$ is the number of nodes in the molecular graph.

We apply global max pooling across nodes to produce a graph-level embedding:

\begin{equation*}
\mathbf{z}_{\text{graph}} = \frac{1}{n} \sum_{i=1}^{n} \mathbf{h}_i^{(2)}.
\end{equation*}

The pooled graph representation $\mathbf{z}_{\text{graph}}$ is then passed through a linear classifier:
\begin{equation*}
\hat{\mathbf{y}}_{\text{graph}} = \text{softmax}(\mathbf{z}_{\text{graph}} \mathbf{W}_g + \mathbf{b}_g),
\end{equation*}
where $\mathbf{W}_g \in \mathbb{R}^{d \times C}$ and $\mathbf{b}_g \in \mathbb{R}^{C}$ are the classification weights and bias, and $C$ is the number of interaction classes.

\subsubsection{Image-only Baseline (ViT)}

The image-only model employs a Vision Transformer (ViT)~\cite{dosovitskiy2021an} to encode 2D chemical structure images.

Given an input image $\mathbf{x}^{\text{img}} \in \mathbb{R}^{H \times W \times 3}$, where $H$ and $W$ denote the height and width, the image is divided into $N$ non-overlapping patches, each of size $P \times P$ pixels.

Each patch is flattened into a vector and linearly projected into a $d$-dimensional embedding space via a learnable matrix $\mathbf{W}_p \in \mathbb{R}^{(P^2 \times 3) \times d}$:
\begin{equation*}
\mathbf{z}_i = \mathbf{x}^{\text{patch}}_i \mathbf{W}_p + \mathbf{b}_p, \quad \forall i = 1, \dots, N
\end{equation*}
where $\mathbf{x}^{\text{patch}}_i$ is the flattened pixel vector of the $i$-th patch.

The sequence of patch embeddings $\{\mathbf{z}_1, \mathbf{z}_2, \dots, \mathbf{z}_N\}$ is prepended with a learnable \texttt{[CLS]} token embedding $\mathbf{z}_{\texttt{[CLS]}} \in \mathbb{R}^d$, and positional encodings are added to preserve spatial information.

The resulting sequence is input into a standard Transformer encoder:
\begin{equation*}
\mathbf{H} = \mathcal{E}_{\text{ViT}} \left(\left[ \mathbf{z}_{\texttt{[CLS]}}, \mathbf{z}_1, \dots, \mathbf{z}_N \right]\right),
\end{equation*}
where $\mathcal{E}_{\text{ViT}}$ denotes the stack of transformer layers.

We extract the output corresponding to the \texttt{[CLS]} token, denoted as $\mathbf{h}_{\texttt{[CLS]}} \in \mathbb{R}^d$, and apply a linear classifier:
\begin{equation*}
\hat{\mathbf{y}}_{\text{image}} = \text{softmax}(\mathbf{h}_{\texttt{[CLS]}} \mathbf{W}_v + \mathbf{b}_v),
\end{equation*}
where $\mathbf{W}_v \in \mathbb{R}^{d \times C}$ and $\mathbf{b}_v \in \mathbb{R}^{C}$ are learnable parameters, $d$ is the hidden dimension, and $C$ is the number of output classes.

\subsection{Multimodal Baselines}

\subsubsection{Late Fusion Baseline}

The late fusion baseline combines predictions from six independent single-modality classifiers, each trained on a distinct representation of drug information. Let $\mathcal{M}$ denote the set of modalities:
\begin{equation*}
\mathcal{M} = \{\text{name}, \text{description}, \text{SMILES}, \text{formula}, \text{graph}, \text{image}\}.
\end{equation*}

Each modality is processed by a dedicated model:
\begin{itemize}[leftmargin=2em]
    \item \textbf{Name, Description, SMILES, Formula:} Each field is input into a separate BioMedBERT encoder to produce four independent textual predictions. For \texttt{formula}, the chemical formula is translated into a sequence of full element names (e.g., \texttt{C20H25N3O} $\rightarrow$ \texttt{carbon 20 hydrogen 25 nitrogen 3 oxygen}) to align with the input of language models.
    \item \textbf{Graph:} The molecular structure graph is encoded using a two-layer GCN.
    \item \textbf{Image:} The 2D chemical structure is processed by a Vision Transformer.
\end{itemize}

Given a drug pair, each model $m \in \mathcal{M}$ produces a predicted label $\hat{y}_m \in \mathcal{C}$, where $\mathcal{C} = \{\texttt{Synergism}, \texttt{Antagonism}, \texttt{New Effect}\}$ is the set of pharmacodynamic interaction classes.

The final prediction $\hat{y}_{\text{late}}$ is obtained through majority voting across modalities:
\begin{equation*}
\hat{y}_{\text{late}} = \argmax_{c \in \mathcal{C}} \sum_{m \in \mathcal{M}} \mathbb{I}(\hat{y}_m = c),
\end{equation*}
where $\mathbb{I}(\cdot)$ is the indicator function.

In the case of a tie (i.e., multiple classes receiving equal votes), we apply a deterministic rule that prioritizes modalities based on their average F1 performance on the MUDI dataset, in the following order:
\texttt{graph} $\rightarrow$ \texttt{name} $\rightarrow$ \texttt{image} $\rightarrow$ \texttt{SMILES} $\rightarrow$ \texttt{formula} $\rightarrow$ \texttt{description}.
This ordering is based on the empirical performance of the individual models on the development set.

\subsubsection{Intermediate Fusion Baseline}

The intermediate fusion baseline constructs a joint representation by integrating six modality-specific embeddings for each drug in the input pair.
For a given drug pair $(d_1, d_2)$, we extract embeddings from the following modalities: name, description, SMILES (text), formula, molecular graph, and chemical image.

Each modality-specific encoder independently processes both drugs:
\begin{equation*}
\begin{aligned}
\mathbf{z}^{(1)}_m &= \mathcal{E}_m(d_1), \\
\mathbf{z}^{(2)}_m &= \mathcal{E}_m(d_2),
\end{aligned}
\quad \text{for each modality } m \in \mathcal{M},
\end{equation*}
where $\mathcal{M}$ is the set of modalities, and $\mathbf{z}^{(i)}_m \in \mathbb{R}^{d_m}$ denotes the embedding of drug $d_i$ in modality $m$.

We concatenate the two drug embeddings for each modality:
\begin{equation*}
\tilde{\mathbf{z}}_m = \left[\mathbf{z}^{(1)}_m;\; \mathbf{z}^{(2)}_m\right] \in \mathbb{R}^{2d_m},
\end{equation*}
and subsequently form the full multimodal representation by concatenating across all modalities:
\begin{equation*}
\mathbf{z}_{\text{fused}} = \left[\tilde{\mathbf{z}}_{\text{name}};\; \tilde{\mathbf{z}}_{\text{desc}};\; \tilde{\mathbf{z}}_{\text{smiles}};\; \tilde{\mathbf{z}}_{\text{formula}};\; \tilde{\mathbf{z}}_{\text{graph}};\; \tilde{\mathbf{z}}_{\text{image}}\right] \in \mathbb{R}^{d_{\text{fused}}},
\end{equation*}
where $d_{\text{fused}} = 2(d_n + d_d + d_s + d_f + d_g + d_i)$.

This joint embedding is passed through a two-layer multilayer perceptron (MLP) with ReLU activation:
\begin{align*}
\mathbf{h}_1 &= \sigma(\mathbf{z}_{\text{fused}} \mathbf{W}_1 + \mathbf{b}_1), \\
\hat{\mathbf{y}}_{\text{inter}} &= \text{softmax}(\mathbf{h}_1 \mathbf{W}_2 + \mathbf{b}_2),
\end{align*}
where:
\begin{itemize}[leftmargin=2em]
    \item $\mathbf{W}_1 \in \mathbb{R}^{d_{\text{fused}} \times d_{\text{hidden}}}$ and $\mathbf{b}_1 \in \mathbb{R}^{d_{\text{hidden}}}$ are parameters of the first MLP layer,
    \item $\mathbf{W}_2 \in \mathbb{R}^{d_{\text{hidden}} \times C}$ and $\mathbf{b}_2 \in \mathbb{R}^{C}$ are parameters of the classification head,
    \item $C$ is the number of pharmacodynamic interaction classes.
\end{itemize}

This fusion strategy enables the model to learn pairwise dependencies and cross-modal interactions between drugs in a unified and expressive representation space.

\subsection{Classification Task Definition}

The central task in MUDI is formulated as a multi-class classification problem over pharmacodynamic drug-drug interactions.
Given a pair of drugs $(d_1, d_2)$, the goal is to predict a single interaction label $y \in \mathcal{C}$ based on their multimodal features.
The label set $\mathcal{C}$ includes four possible classes:
\begin{itemize}[leftmargin=2em]
    \item \texttt{Synergism} -- drug $d_1$ enhances the effect of $d_2$.
    \item \texttt{Antagonism} -- drug $d_1$ reduces or nullifies the effect of $d_2$.
    \item \texttt{New Effect} -- the combination produces a novel outcome not present in individual use.
    \item \texttt{No Interaction} — no significant pharmacodynamic interaction is known or observed.
\end{itemize}

Each sample is annotated with one of the four mutually exclusive labels, with directional semantics included for \texttt{Synergism} and \texttt{Antagonism}.
Specifically, $(d_1, d_2)$ and $(d_2, d_1)$ may correspond to different labels or directions unless symmetry is explicitly annotated (e.g., in \texttt{New Effect} cases).

To align with clinical interest in detecting meaningful drug interactions, our evaluation protocol concentrates on the three \textbf{positive interaction classes} -- \texttt{Synergism}, \texttt{Antagonism}, and \texttt{New Effect}.
The \texttt{No Interaction} label is used during training to simulate realistic class imbalance and improve discrimination, but is excluded from performance metric computation during test-time evaluation, following best practices in biomedical literature~\cite{peng2018extracting}.


\subsection{Prediction Thresholds}

All models produce a probability distribution over the four interaction classes via a softmax output layer.  
Final class predictions are made using a maximum likelihood decision rule:
\begin{equation*}
\hat{y} = \argmax_{c \in \mathcal{C}} \; p(c \mid \mathbf{x}),
\end{equation*}
where $\mathcal{C} = \{\texttt{Synergism}, \texttt{Antagonism}, \texttt{New Effect}, \texttt{No Interaction}\}$, and $p(c \mid \mathbf{x})$ is the predicted probability for class $c$ given multimodal input $\mathbf{x}$.

No additional confidence thresholding is applied. This choice ensures fair and consistent comparison across models, particularly under class imbalance conditions.

\subsection{Reproducibility}

To promote transparency and facilitate fair comparison, we standardize all experimental procedures as follows:
\begin{itemize}[leftmargin=2em]
    \item \textbf{Randomness control:} All experiments are conducted with fixed random seeds for \texttt{PyTorch}, \texttt{NumPy}, and system-level generators to ensure consistent results across runs.
    \item \textbf{Dataset splits:} We use the same predefined training and test sets for all baseline models and fusion strategies.
    \item \textbf{Evaluation consistency:} All models are evaluated using a unified set of metrics and evaluation scripts, ensuring consistent treatment of prediction outputs under both direction-aware and direction-agnostic settings.
\end{itemize}

The full evaluation pipeline, including scoring functions and matching logic, is publicly available in the official repository (see Appendix~\ref{appendix:access}). This setup enables full replication of our results and supports future benchmarking efforts on the MUDI dataset.

\section{Evaluation Protocols and Settings}
\label{appendix:evaluation}

\subsection{Evaluation Metrics}

To evaluate model performance on clinically meaningful interaction types, we report standard classification metrics computed over the three positive classes: \texttt{Synergism}, \texttt{Antagonism}, and \texttt{New Effect}.

\paragraph{Precision (P)}  
For each class, Precision is the proportion of correctly predicted instances among all instances assigned to that class:
\begin{equation*}
\text{Precision}_c = \frac{\text{TP}_c}{\text{TP}_c + \text{FP}_c},
\end{equation*}
where $\text{TP}_c$ and $\text{FP}_c$ denote true positives and false positives for class $c$.

\paragraph{Recall (R)}  
Recall is the proportion of correctly predicted instances among all actual instances of the class:
\begin{equation*}
\text{Recall}_c = \frac{\text{TP}_c}{\text{TP}_c + \text{FN}_c},
\end{equation*}
where $\text{FN}_c$ is the number of false negatives for class $c$.

\paragraph{F1 Score (F1)}  
The F1 score is the harmonic mean of Precision and Recall:
\begin{equation*}
\text{F1}_c = \frac{2 \cdot \text{Precision}_c \cdot \text{Recall}_c}{\text{Precision}_c + \text{Recall}_c}.
\end{equation*}

\paragraph{Micro-averaged Metrics}  
Micro-averaging aggregates true positives, false positives, and false negatives across all positive classes before computing Precision, Recall, and F1:
\begin{equation*}
\text{Precision}_{\text{micro}} = \frac{\sum_c \text{TP}_c}{\sum_c (\text{TP}_c + \text{FP}_c)}, \quad
\text{Recall}_{\text{micro}} = \frac{\sum_c \text{TP}_c}{\sum_c (\text{TP}_c + \text{FN}_c)},
\end{equation*}
\begin{equation*}
\text{Micro-F1} = \frac{2 \cdot \text{Precision}_{\text{micro}} \cdot \text{Recall}_{\text{micro}}}{\text{Precision}_{\text{micro}} + \text{Recall}_{\text{micro}}}.
\end{equation*}

\paragraph{Macro-averaged Metrics}  
Macro-averaging computes the unweighted mean of the per-class metrics. We first compute macro-averaged Precision and Recall:
\begin{equation*}
\text{Precision}_{\text{macro}} = \frac{1}{3} \sum_{c=1}^{3} \text{Precision}_c, \quad
\text{Recall}_{\text{macro}} = \frac{1}{3} \sum_{c=1}^{3} \text{Recall}_c.
\end{equation*}
Then, we define the Macro-F1 score as the harmonic mean of these macro-averaged values:
\begin{equation*}
\text{Macro-F1} = \frac{2 \cdot \text{Precision}_{\text{macro}} \cdot \text{Recall}_{\text{macro}}}{\text{Precision}_{\text{macro}} + \text{Recall}_{\text{macro}}}.
\end{equation*}

\paragraph{Treatment of Negative Class (\texttt{No Interaction})}  
Although the negative class is included during training to enhance model calibration and decision boundaries, it is excluded from all evaluation metrics.
This decision reflects established practice in biomedical relation extraction~\cite{peng2018extracting}, which emphasizes performance on clinically actionable positive interactions.

\subsection{Evaluation Settings}

We evaluate model performance under two distinct matching settings to reflect different use scenarios:

\begin{itemize}[leftmargin=2em]
    \item \textbf{Direction-aware Matching.}
    Drug pairs are treated as ordered tuples; that is, (\texttt{DRUG1}, \texttt{DRUG2}) and (\texttt{DRUG2}, \texttt{DRUG1}) are considered distinct.
    This setting requires models to not only detect the correct interaction type but also capture the directionality -- i.e., which drug initiates or modulates the effect.
    
    \item \textbf{Direction-agnostic Matching.}
    Drug pairs are treated as unordered sets. A prediction is considered correct if it matches the ground-truth interaction type for either (\texttt{DRUG1}, \texttt{DRUG2}) or its reversed pair (\texttt{DRUG2}, \texttt{DRUG1}).
    This relaxed setting reflects clinical cases where directionality is either symmetric or not explicitly defined.
\end{itemize}

Unless otherwise noted, all results reported in the main text follow the stricter direction-aware setting, which better aligns with real-world pharmacodynamic modeling.

\section{Training Environment and Hyperparameter Configurations}
\label{appendix:environment_hyperparams}

This appendix details the computational environment, software stack, and hyperparameter configurations used for training and evaluating all baseline models.

\subsection{Hardware and Software Environment}

Experiments are conducted on a Linux server with the following specifications:
\begin{itemize}[leftmargin=2em]
    \item CPU: Intel(R) Xeon(R) CPU (2.2 GHz, 2 cores)
    \item GPU: 2$\times$ NVIDIA T4 GPUs (16GB VRAM each)
    \item RAM: 32GB DDR4 Memory
    \item Storage: 128GB NVMe SSD
\end{itemize}
The software environment is standardized as follows:
\begin{itemize}[leftmargin=2em]
    \item Operating System: Ubuntu 22.04 LTS
    \item Python: 3.10
    \item PyTorch: 2.0.1
    \item CUDA: 11.8
    \item Transformers Library (HuggingFace): 4.31
    \item DGL (Deep Graph Library): 1.1.1
    \item scikit-learn: 1.2.2
    \item RDKit: 2022.09.5
    \item Additional packages: NumPy 1.24, SciPy 1.10, Matplotlib 3.7
\end{itemize}
%

\subsection{General Training Settings}

Unless otherwise specified, the following settings are shared across all baseline models:
\begin{itemize}[leftmargin=2em]
    \item Optimizer: Adam~\cite{kingma2015adam}
    \item Initial learning rate: $5\times10^{-5}$
    \item Batch size: 32
    \item Learning rate scheduler: linear decay with warm-up (10\% of total steps)
    \item Weight decay: $1\times10^{-2}$
    \item Dropout rate: 0.1 (applied after embeddings and in MLPs)
    \item Number of epochs: 100
    \item Gradient clipping: maximum norm of 1.0
\end{itemize}
Early stopping is applied based on validation loss with a patience of 5 epochs.

\subsection{Model-Specific Hyper-parameters}

\subsubsection{Text-only Baseline (BioMedBERT)}
\begin{itemize}[leftmargin=2em]
    \item Pretrained checkpoint: `BioMedBERT-Base (uncased)'
    \item Maximum sequence length: 512 tokens
    \item Hidden size: 768
    \item Number of transformer layers: 12
    \item Number of attention heads: 12
    \item Fine-tuned end-to-end on MUDI dataset
\end{itemize}

\subsubsection{Graph-only Baseline (GCN)}
\begin{itemize}[leftmargin=2em]
    \item Number of GCN layers: 2
    \item Hidden dimension ($d_{\text{hidden}}$): 768
    \item Input features: 37 atom features (one-hot encoded)
    \item Activation: ReLU
    \item Readout: global max pooling
\end{itemize}

\subsubsection{Image-only Baseline (ViT)}
\begin{itemize}[leftmargin=2em]
    \item Pretrained checkpoint: `ViT-B/16'
    \item Image size: 1000$\times$800 pixels
    \item Patch size: 16$\times$16
    \item Hidden size: 768
    \item Number of transformer layers: 12
    \item Number of attention heads: 12
    \item MLP head dimension: 3072
    \item Fine-tuned end-to-end on MUDI dataset
\end{itemize}

\subsubsection{Late Fusion Baseline}
\begin{itemize}[leftmargin=2em]
    \item No additional training is performed.
    \item Predictions are aggregated from six single-modality models: Name, Description, SMILES, Formula, Graph, and Image.
    \item Tie-breaking priority: \texttt{graph} $\rightarrow$ \texttt{name} $\rightarrow$ \texttt{image} $\rightarrow$ \texttt{SMILES} $\rightarrow$ \texttt{formula} $\rightarrow$ \texttt{description}.
\end{itemize}

\subsubsection{Intermediate Fusion Baseline}
\begin{itemize}[leftmargin=2em]
    \item Embedding dimension for each modality: $d = 768$.
    \item Concatenated embedding dimension: $d_{\text{fusion}} = 6 \times 768 = 4608$.
    \item Fusion MLP: Two fully connected layers.
    \item Hidden dimension: 1024.
    \item Activation: ReLU.
    \item Dropout: 0.1 after each layer.
    \item Output: 4-way softmax classification.
\end{itemize}

\subsection{Training Time}

On average, training our baseline model takes:

\begin{itemize}[leftmargin=2em]
    \item \textbf{Single-modality model}: 3.5--4 hours per modality (BioMedBERT, GCN, ViT, etc.).
    \item \textbf{Intermediate Fusion}: 4--4.5 hours, including preloading all modality-specific encoders and training the fusion MLP.
    \item \textbf{Late Fusion}: No additional training time, as predictions are directly aggregated from pretrained single-modality models.
\end{itemize}

All timing estimates are based on dual-GPU training using two NVIDIA T4 GPUs with 16GB memory on each GPU.

\section{Additional Results}
\subsection{Single-Modality Analysis}
\label{appendix:singlemodal}

Table~\ref{tab:single-modal-result} presents the classification performance of each individual modality on the MUDI dataset under both direction-aware and direction-agnostic matching.
Among all modalities, the molecular \textbf{graph}-based model performs best, achieving a Micro-F1 of 65.44\% and Macro-F1 of 57.36\% in the direction-agnostic setting.
This demonstrates the effectiveness of topological molecular representations in capturing pharmacodynamic interactions.
The \textbf{name}-based model also yields surprisingly strong results, suggesting that drug identity alone encodes useful priors, especially when paired drugs have known interaction profiles.

In contrast, modalities like \textbf{SMILES}, \textbf{formula}, and \textbf{description} show limited standalone performance, particularly for the \textit{New Effect} class.
This result is consistent with the difficulty of extracting discriminative features from raw strings or sparse chemical formulas, and the noisiness of unstructured textual fields.
Direction-aware results are consistently lower across all modalities, highlighting the increased challenge when models must account for interaction asymmetry.

\begin{table*}[!t]
\caption{Results of Single-Modality baselines on our MUDI dataset.}
\label{tab:single-modal-result}
\resizebox{\textwidth}{!}{%
\begin{tabular}{lc|ccccc|ccccc}
\hline
\multicolumn{2}{l|}{} & \multicolumn{5}{c|}{\textbf{Direction-aware Matching}} & \multicolumn{5}{c}{\textbf{Direction-agnostic Matching}} \\ \hline
\multicolumn{1}{l|}{Modality} & Metric & Synergism & Antagonism & New Effect & Macro-averaged & Micro-averaged & Synergism & Antagonism & New Effect & Macro-averaged & Micro-averaged \\ \hline
\multicolumn{1}{l|}{\multirow{3}{*}{Name}} & Precision & 38.73 & 41.13 & 33.84 & 37.90 & 38.81 & 62.92 & 59.21 & 53.74 & 58.62 & 62.25 \\
\multicolumn{1}{l|}{} & Recall & 60.57 & 36.03 & 18.31 & 38.30 & 53.53 & 68.21 & 36.03 & 35.1 & 46.45 & 61.18 \\
\multicolumn{1}{l|}{} & F1 & 47.25 & 38.41 & 23.76 & 38.10 & 44.99 & 65.46 & 44.8 & 42.46 & 51.83 & 61.71 \\ \hline
\multicolumn{1}{l|}{\multirow{3}{*}{Description}} & Precision & 41.86 & 0.00 & 0.00 & 13.95 & 41.86 & 65.78 & 0.00 & 0.00 & 21.93 & 65.78 \\
\multicolumn{1}{l|}{} & Recall & 65.33 & 0.00 & 0.00 & 21.78 & 50.03 & 71.92 & 0.00 & 0.00 & 23.97 & 56.31 \\
\multicolumn{1}{l|}{} & F1 & 51.02 & 0.00 & 0.00 & 17.01 & 45.58 & 68.72 & 0.00 & 0.00 & 22.90 & 60.68 \\ \hline
\multicolumn{1}{l|}{\multirow{3}{*}{SMILES}} & Precision & 32.35 & 30.64 & 0.00 & 21.00 & 32.17 & 54.32 & 47.95 & 0.00 & 34.09 & 53.58 \\
\multicolumn{1}{l|}{} & Recall & 62.27 & 36.76 & 0.00 & 33.01 & 53.33 & 68.93 & 36.76 & 0.00 & 35.23 & 60.24 \\
\multicolumn{1}{l|}{} & F1 & 42.58 & 33.42 & 0.00 & 25.67 & 40.13 & 60.76 & 41.62 & 0.00 & 34.65 & 56.72 \\ \hline
\multicolumn{1}{l|}{\multirow{3}{*}{Formula}} & Precision & 29.75 & 21.11 & 24.1 & 24.99 & 28.56 & 50.68 & 35.78 & 36.94 & 41.13 & 48.51 \\
\multicolumn{1}{l|}{} & Recall & 56.55 & 31.57 & 8.07 & 32.06 & 48.98 & 63.99 & 31.57 & 15.84 & 37.13 & 56.22 \\
\multicolumn{1}{l|}{} & F1 & 38.99 & 25.3 & 12.09 & 28.09 & 36.08 & 56.56 & 33.54 & 22.17 & 39.03 & 52.08 \\ \hline
\multicolumn{1}{l|}{\multirow{3}{*}{Image}} & Precision & 32.37 & 28.77 & 29.54 & 30.23 & 31.86 & 54.45 & 45.15 & 47.94 & 49.18 & 53.05 \\
\multicolumn{1}{l|}{} & Recall & 54.82 & 37.94 & 12.76 & 35.17 & 48.96 & 62.09 & 37.94 & 25.01 & 41.68 & 56.25 \\
\multicolumn{1}{l|}{} & F1 & 40.71 & 32.72 & 17.82 & 32.51 & 38.6 & 58.02 & 41.23 & 32.87 & 45.12 & 54.6 \\ \hline
\multicolumn{1}{l|}{\multirow{3}{*}{Graph}} & Precision & 53.09 & 61.27 & 52.30 & 55.55 & 54.05 & 77.23 & 74.56 & 71.24 & 74.34 & 76.69 \\
\multicolumn{1}{l|}{} & Recall & 53.96 & 46.09 & 17.79 & 39.28 & 49.91 & 61.25 & 43.88 & 34.97 & 46.70 & 57.07 \\
\multicolumn{1}{l|}{} & F1 & 53.52 & 52.61 & 26.55 & 46.02 & 51.90 & 68.32 & 55.25 & 46.92 & 57.36 & 65.44 \\ \hline
\end{tabular}%
}
\end{table*}

\begin{figure}[!t]
    \centering
    \includegraphics[width=1\linewidth]{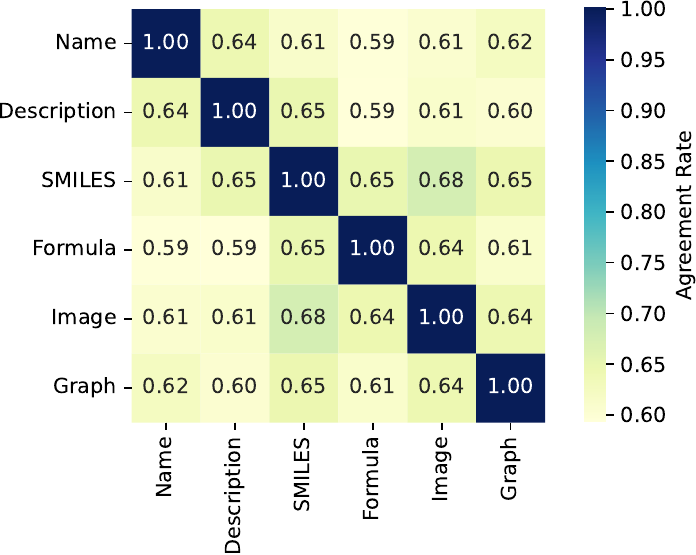}
    \caption{Agreement rate between different modalities.}
    \label{fig:modality_agreement}
\end{figure}

To further understand how different modalities contribute to predictions, we visualize their agreement in Figure~\ref{fig:modality_agreement}.
The heatmap shows that while all modality pairs exhibit moderate correlation (typically between 0.59 and 0.68), \textbf{SMILES} and \textbf{image} achieve the highest agreement at 0.68, likely due to shared molecular-level information.
This finding motivates future work on modality selection, weighted ensembling, or modality-specific gating to optimize fusion strategies.

\subsection{Additional Ablation Studies}

\begin{figure}
    \centering
    \includegraphics[width=\linewidth]{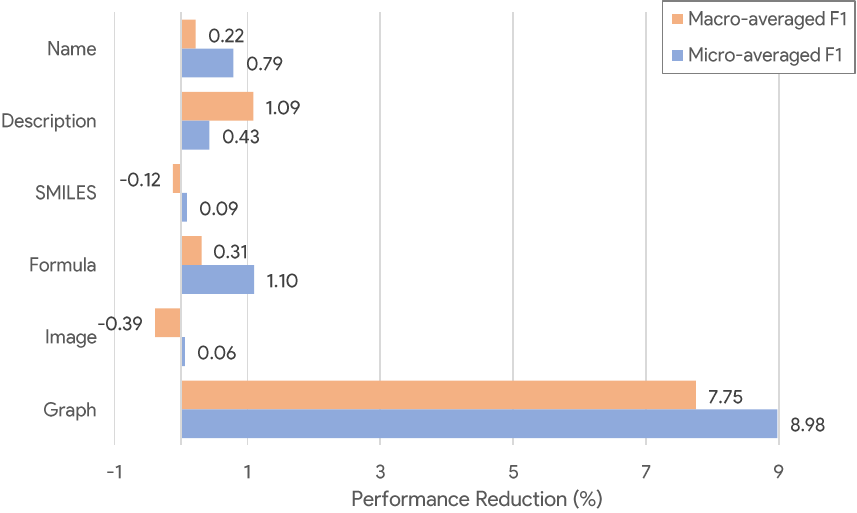}
    \caption{Macro and Micro F1 reduction with Intermediate Fusion.} 
    \label{fig:ablation-result-intermediate}
\end{figure}

\begin{figure}
    \centering
    \includegraphics[width=\linewidth]{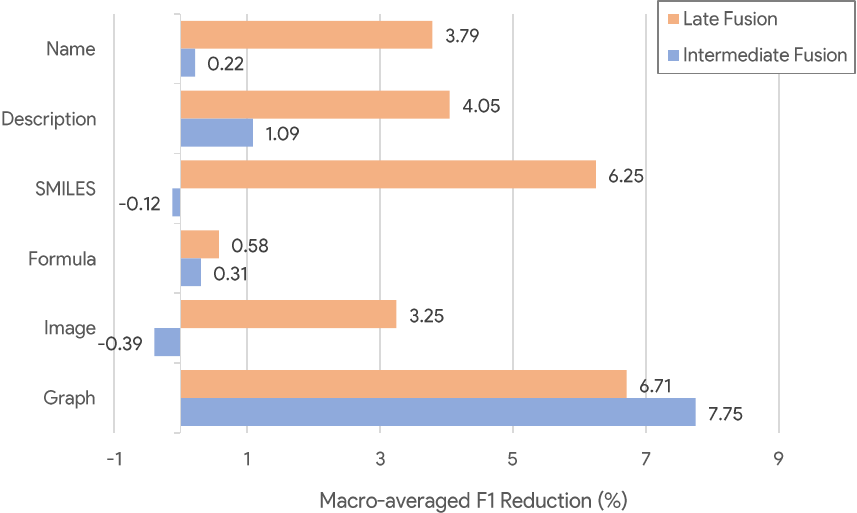}
    \caption{Macro F1 reduction between two fusion strategies.} 
    \label{fig:ablation-result-macro-two-fusion}
\end{figure}

To further understand the contribution of modalities, we conduct several ablation experiments, each removing a modality to exhibit its impact on performance. Figure \ref{fig:ablation-result-intermediate} illustrates the reductions in macro and micro F1 scores for Intermediate Fusion. Additionally, Figure \ref{fig:ablation-result-macro-two-fusion} shows the decrease in the macro-averaged F1 metric across two fusion strategies.

The ablation study results in Figure \ref{fig:ablation-result-intermediate} reveal a significant impact from Molecular Graph and demonstrate its enormous impact, as the scores are reduced by around 9\%. Conversely, the performance when excluding Image and SMILE channels slightly improves the micro-averaged F1 but declines the macro-averaged F1, indicating potential conflict when adding these input sources. All other modalities contribute positively to prediction accuracy, however their individual impact is comparatively small.

As shown in Figure \ref{fig:ablation-result-macro-two-fusion}, Late Fusion generally shows a greater performance reduction than Intermediate Fusion when a specific modality is ablated. While removing SMILES in Intermediate Fusion shows a minor increase, its exclusion in Late Fusion results in a substantial 6.25\% reduction, which underscores the inherent importance of this modality. This highlights the need for advanced fusion methods that can effectively integrate heterogeneous information from diverse modalities, as current baseline models may not fully leverage their complementary contributions.

\end{document}